\documentclass[conference]{IEEEtran}
\IEEEoverridecommandlockouts
\usepackage{cite}
\usepackage{amsmath,amssymb,amsfonts}
\usepackage{algorithm}
\usepackage{algorithmic}
\usepackage{graphicx}
\usepackage{textcomp}
\usepackage{xcolor}
\def\BibTeX{{\rm B\kern-.05em{\sc i\kern-.025em b}\kern-.08em
    T\kern-.1667em\lower.7ex\hbox{E}\kern-.125emX}}

\usepackage{multirow} 
\definecolor{deepblue}{rgb}{0, 0.0, 0.85}
\definecolor{deepgreen}{rgb}{0, 0.5, 0}

\definecolor{deeppurple}{rgb}{0.4, 0, 0.5}

\definecolor{deepyellow}{rgb}{0.8, 0.8, 0}

\definecolor{skyblue}{rgb}{0, 0.7, 0.85}

\newcommand{\myrefmark}[1]{\textsuperscript{#1}}

\begin{document}

\title{Video Flow as Time Series: Discovering Temporal Consistency and Variability for VideoQA

\thanks{\hrule This work was supported by the NSFC NO. 62172138, 62202139 and U23B2031. This work was also partially supported by the Fundamental Research Funds for the Central Universities NO. JZ2024HGTG0310.
\\ \textbf{This work has been accepted in ICME2025}}
}

\author{
\IEEEauthorblockN{
Zijie~Song\myrefmark{1},
Zhenzhen~Hu\myrefmark{1},
Yixiao~Ma\myrefmark{2},
Jia~Li\myrefmark{1} and
Richang Hong\myrefmark{1}}
\IEEEauthorblockA{\myrefmark{1}Hefei University of Technology, Hefei, China \\
\myrefmark{2}University of Science and Technology of China, Hefei, China\\
zjsonghfut@gmail.com; huzhen.ice@gmail.com; mayx@mail.ustc.edu.cn; jiali@hfut.edu.cn; hongrc.hfut@gmail.com}
}

\maketitle

\begin{abstract}
Video Question Answering (VideoQA) is a complex video-language task that demands a sophisticated understanding of both visual content and temporal dynamics. Traditional Transformer-style architectures, while effective in integrating multimodal data, often simplify temporal dynamics through positional encoding and fail to capture non-linear interactions within video sequences. In this paper, we introduce the Temporal Trio Transformer (T3T), a novel architecture that models time consistency and time variability. The T3T integrates three key components: Temporal Smoothing (TS), Temporal Difference (TD), and Temporal Fusion (TF). The TS module employs Brownian Bridge for capturing smooth, continuous temporal transitions, while the TD module identifies and encodes significant temporal variations and abrupt changes within the video content. Subsequently, the TF module synthesizes these temporal features with textual cues, facilitating a deeper contextual understanding and response accuracy.
The efficacy of the T3T is demonstrated through extensive testing on multiple VideoQA benchmark datasets. Our results underscore the importance of a nuanced approach to temporal modeling in improving the accuracy and depth of video-based question answering.

\end{abstract}


\vspace{-8pt}

\section{Introduction}
\label{sec:intro}
\begin{figure}[t]
\centering
\includegraphics[width=1.0\columnwidth]{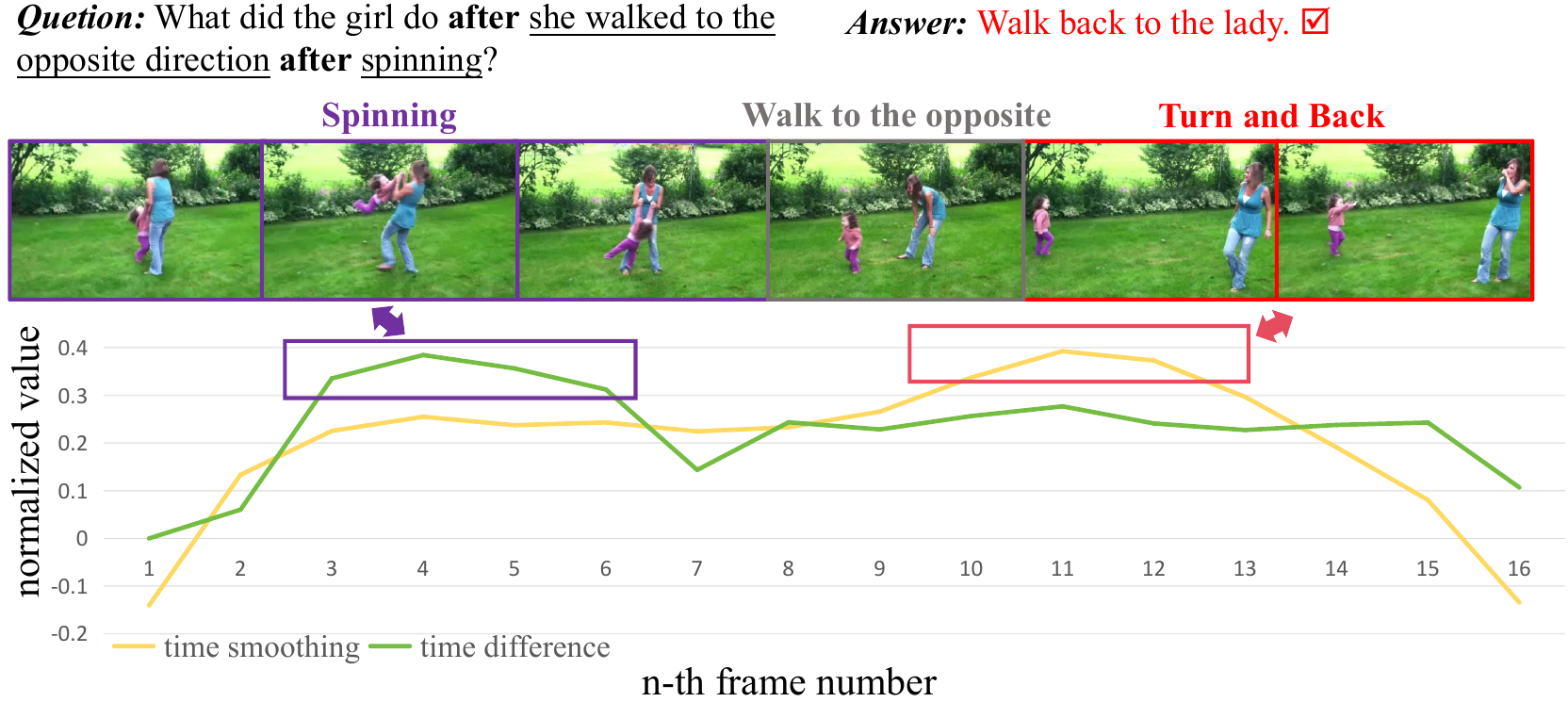} 
\vspace{-10pt}
\caption{The visualization comparison of normalized time smoothing and time difference features. Given the question, they focus on the different parts of video. The consistency learned from TS provides high values on the frames related to the ‘Turn and Back’. The variability extracted from TD pays close attention to the local feature change where ‘Spinning’ is drastic action.}
\label{fig:example}
\vspace{-20pt}
\end{figure}

\vspace{-3pt}
In the realm of video-language tasks, Video Question Answering (VideoQA) stands out as one of the challenges that demand a high degree of temporal understanding where video and language are both sequential forms of information characterized by their temporality. 
This task requires models not only to process visual content but also to reason across the temporal sequence of events in a video in response to specific questions~\cite{zhong2022video,li2023discovering,gao2023mist,yu2024self}. The intricacies of time consistency and time variability within these sequences pose a significant challenge. 
As shown in Fig.~\ref{fig:example}, the question involves a complex temporal reasoning, as models must accurately interpret both the continuous flow and abrupt changes in content to provide coherent and contextually relevant answers.

Transformer-style architectures for VideoQA~\cite{xiao2022video,zong2024video} have excelled particularly in integrating complex, multi-modal data streams, thereby enriching the interpretative depth of video content. 
However, their reliance on positional encoding to capture time sequences often results in a linear and oversimplified representation of temporal dynamics, limiting the potential for a deeper, more accurate temporal analysis. Consequently, there remains a pressing need for innovations that can more effectively model the intricate temporal relationships inherent in video data.

Video flow can be conceptualized as a sequence of time series, where each frame represents a discrete point in time, contributing to a continuous narrative. This perspective allows for the application of time series analysis techniques to effectively capture and interpret the dynamic temporal patterns inherent in video data. Modeling video flow as the sequence of time series has been preliminarily attempted to learn the temporal representations~\cite{danghierarchical,sermanet2018time,chen2021rspnet}. 
With the finding that evolution process of continuous dynamic frames should follow the coherent constraint, \cite{zhang2023modeling} model video as a stochastic process to enforce an arbitrary frame to agree with a time-variant Gaussian distribution. However, these methods have only modeled partial aspects of the temporal nature of video, or have relied on overly complex assumptions and constraints without modeling the most essential and general temporal variation, highlighting the need for a more balanced approach that effectively integrates both time consistency and time variability.

In this paper, we combine temporal perspectives from time consistency and time variability in a comprehensive modeling with \textit{Temporal Smoothing} (TS), \textit{Temporal Difference} (TD) and \textit{Temporal Fusion} (TF) and propose a novel architecture \textbf{T}emporal \textbf{T}rio \textbf{T}ransformer (T3T). 
The TS module utilizes Brownian Bridge to ensure smooth transitions across frames, capturing the overarching narrative flow. 
Meanwhile, the TD module focuses on identifying and encoding significant changes and events, allowing the model to respond to dynamic shifts within the video. 
TS and TD not only captures the consistency of smoothing temporal features across time but also highlights their variability. As shown in Fig.~\ref{fig:example}, TS and TD modules work collaboratively to trace the temporal clues from the entire video. 
Sequentially, the TF module integrates these processed streams, aligning and synthesizing them with textual cues to produce contextually relevant and temporally coherent responses.
Finally, both self-attention and cross-attention mechanisms are set to refine and predict the final answers with T3T output in Answer Prediction.
Extensive experiments conducted on multiple VideoQA datasets have demonstrated that superior capabilities of T3T in handling complex temporal dynamics and recognizing significant local changes within videos.

The summary of our contributions as follow:

\begin{itemize}
\item We discover and analyze the prominent time clues and the potential of treating video flow as time series to uncover essential characteristics. Through Brownian Bridge and Difference on video features, we provide comprehensive and interpretable temporal modeling for VideoQA.
\item We introduce a novel architecture, the T3T, which effectively models both time consistency and time variability in video. This architecture consists of three key components: TS, TD and TF, each designed to capture different aspects of temporal dynamics.
\item The experimental results demonstrate significant improvements in handling complex temporal dynamics and reasoning. The validation underscores our model's accuracy in interpreting and answering video-based questions.

\end{itemize}

\section{Related work}

\subsection{Video Question Answering}

VideoQA leverages comprehensive video analysis and understanding to facilitate more difficult and complicated question answering in videos through both visual and textual information of its content with temporal extension. 
Graph-based inference methods~\cite{wang2021dualvgr,park2021bridge,liu2022cross} have demonstrated superior performance by explicitly capturing the object-level representations, their relationships, and dynamics to involve encoding the video. 
Self-supervised pretraining models with transformer-style architecture~\cite{selva2023video,peng2023efficient,li2023transformer} and methods~\cite{yu2024self} combined Large Language Models make significant advancements to address the demands of reasoning and understanding. However, these methods almost establish spatial and temporal relationships through model attention or positional encoding without taking the pure temporal modeling into consideration where video distinguished from other visual-text tasks contain temporal cues and sequential information. In this paper, we explore the most fundamental and general temporal characteristic of video to integrate both time consistency and time variability.

\subsection{Temporal Learning in Video Analysis}

The study of temporal learning in video analysis has been a focal point aiming to capture the flow and evolution of video events. Different from capturing the sequential nature~\cite{yu2018joint,fan2019heterogeneous,jiang2020divide}, the view~\cite{xiao2022aaaivideo} argued that video is presented in sequence, but visual elements are hierarchical in semantic space. An end-to-end pipeline~\cite{buch2022revisiting} was proposed to select only a single frame as input for downstream tasks which still keeps a bright performance. Similarly, frame selection methods~\cite{hu2023dual,li2023discovering} based on video segments focused on choosing the most important few frames. The key frames themselves have no temporal characteristics and require several complicated modules and steps to complete. Exploring the temporal characteristic of video, the method~\cite{zhang2023modeling} approximated video as stochastic processes and utilized sequence contrastive learning to pretrain video representation. Inspired by this view which allows the potential for video to degenerate into time series, we introduce TS to addresses temporal evolution and TD to retain significant variations to model temporal structure for VideoQA task.

\section{Methods}

\begin{figure*}[!ht]
\centering
\includegraphics[width=0.7\textwidth]{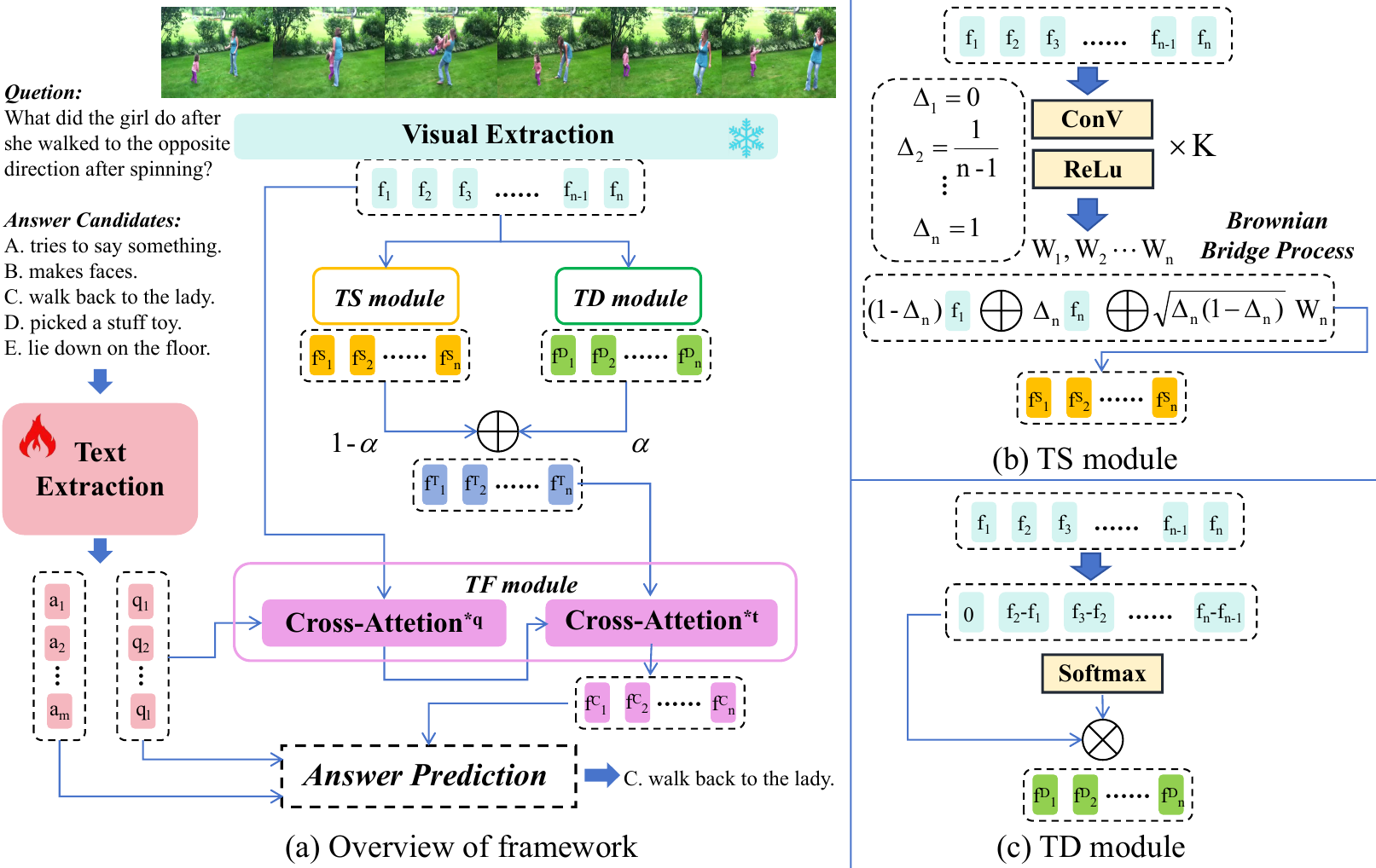} 
\vspace{-5pt}
\caption{Overview of our framework for VideoQA. Fig.~\ref{framework}~(a) show the entire process begins with video frames and textual data are separately encoded. 
The Temporal Trio Transformer (T3T) incorporates three modules: Temporal Smoothing (TS), Temporal Difference (TD), and Temporal Fusion (TF), to deeply capture the temporal dynamics within the video. TS module smooths temporal transitions using the Brownian Bridge Process, detailed in Fig.~\ref{framework}~(b). TD module captures abrupt temporal variations through difference operations, detailed in Fig.~\ref{framework}~(c). These features via the balance value $\alpha$ are fused with textual information in the TF. Finally, The Answer Prediction stage integrates and refines these multimodal features enabling accurate answer selection.}
\label{framework}
\vspace{-15pt}
\end{figure*}

Our framework is shown in Fig.~\ref{framework} to implement VideoQA task
including three parts: Visual-text Representation Extraction, Temporal Trio Transformer, and Answer Prediction. 


\subsection{Visual-text Representation Extraction}

For given the video $v$ and correlative question $q$, VideoQA task allows model to predict the answer $\hat{a}$ from the answer candidates set $\mathbf{A}$. We uniformly sample N video frames to represent the entire video event and extract frames set $\{f_n\}_{1:N} \in \mathbb{R}^{N \times D}$ via a frozen pretrained visual model during the training where the feature is in $D$-dimensional space. For text, the question $q$ and the answer candidates set $\mathbf{A}$ should be considered separately via a pretrained language model which requires fine-tuned in training and has the same feature dimension $D$. $q$ is encoded as word embedding sequence $\{q_l\}_{1:L} \in \mathbb{R}^{L \times D}$ with $L$ tokens. The answer candidates are encoded as $\{a_m\}_{1:M} \in \mathbb{R}^{M \times D}$ by number of candidates $M$.

\subsection{Temporal Trio Transformer (T3T)}
T3T has three modules Temporal Smoothing (TS), Temporal Difference (TD) and Temporal Fusion (TF) to model the representation of video. Video features are highly redundant and have a strong degree of degradation during the transition to answer classification. It provides the possibility of modeling from a sequential time series perspective. \textbf{\textit{TS}} is proposed to generate time series by stochastic processing and obtains smooth visual representation displaying the explicit video type and event via smooth change between continuous frames of a video. However, this process inevitably results in the loss of information where significant local action and scene changes in the video flow are urgently required. A simply and effectively \textbf{\textit{TD}} proposed aims to preserve and remember the local features by value changes. Two modules conjointly complete the temporal modeling of video features. Given initial video features $\{f_n\}_{1:N}$ as inputs, the $\{f^S_n\}_{1:N}$ from TS and the $\{f^D_n\}_{1:N}$ from TD are combined via a hyperparameter $\alpha$ to adjust the sensitivity of the dataset with two types of features. The output of $n\text{-}th$ frame feature $f^T_n \in \{f^T_n\}_{1:N}$ is as follow: 
\begin{equation}
    f^T_n = ( 1-\alpha ) f^S_n + \alpha f^D_n,
\end{equation}
where the balance value $\alpha$ establishes the proportion between temporal smoothness and local discontinuity to achieve stable modeling. 
Then, \textbf{\textit{TF}} is proposed as reasoning structure to introduce the question information and force the model to focus more on prominent and relevant video content where multimodal interactions are involved. We employed the design of shared parameter and cross-attention architecture to obtain the final video feature representation $f^C_n \in \{f^C_n\}_{1:N}$. 

\subsubsection{Temporal Smoothing (TS)}

Capturing the inherent smoothness and continuity of video representation is crucial for video analysis and understanding. The inherent flow and evolution of video events cannot be fully explained by merely enforcing spatial-temporal correlations through attention and position-aware representations. To address this issue, we propose TS nodule integrating video flow as time series with Brownian Bridge Process to capture the dynamic evolution of the subtle nuances and transient behaviors that may be obscured by deterministic formulations.


Assume that the video flow can approximate as a Brownian Bridge Process and define the smoothing feature $f^S_n \in \{f^S_n\}_{1:N}$ as follow: 
\begin{equation}
    f^S_n = (1-\Delta_n) f_1 + \Delta_n f_N + \sqrt{\Delta_n(1-\Delta_n)}W_n,
\end{equation}
\begin{equation}
    W_n = ConV_K(f_n), f_n \in \{f_n\}_{1:N},
\end{equation}
\begin{equation}
   ConV_K(\cdot) = [nn.Conv1d(\cdot), nn.ReLU(\cdot)]_K,
\end{equation}
where the Brown Bridge process satisfies $f^S_1 = f_1$ and $f^S_N = f_N$. $\{\Delta_n\}_{1:N}$ is the time step with a uniform distribution from $0$ to $1$. $W_n$ is the stochastic element generated by initial $f_n$ and we use $ConV_K(\cdot)$ with K Conv and ReLu layers to learn this process.

The start and end of the bridges are uniquely determined by the initial video frame features, which enable the modeling process to take into account both the learned video content and the uncertainties during dynamics. Through the modeling of the stochastic process, video frame feature $f^S_n$ has closer temporal distance which is more similar with a smaller change range and consistent with the dynamic changes of video flow, continuous and smooth.

\subsubsection{Temporal Difference (TD)} 

Modeling based on stochastic processes treats the video flow as time series, which offers inherent continuity and smoothness of the temporal dynamic representation. However, it may struggle to faithfully capture any abrupt and significant local changes due to frame samples. Thus, we propose TD module to extract and preserve such significant step signals by simple and effective frame difference. The difference features make it easier to remember changes like local actions and scene switches.
We define the $n\text{-}th$ frame variability feature by difference operation and set the interval $I$ to decide on the span of the difference as follow:
\begin{equation}
    f^D_n = (f_n - f_{n-1-I})Softmax(f_n - f_{n-1-I}),
\end{equation}
where $f^D_n = 0$ when $n \leq I$. Softmax function further enhances and emphasizes intensity of the discontinuous representation by probabilistic amplitude. Intuitively, interval I should not be too large in order to preserve the changes between neighboring frames where the sampled frames have already a time span.

The difference operation provides a direct way without trainable parameters to capture the rate of change between video frames. This complements the continuous and smooth nature of the TS used to model the overall video flow.

\subsubsection{Temporal Fusion (TF)}
TF aims to address two key issues in temporal modeling VideoQA task. One is introducing question as text-guided information to video representation, which can allow the model to focus on the most relevant and salient aspects of the video content. Question as text input determines which parts of the video the model should focus on in order to find the answer. $\{f_n\}_{1:N}$ as query, $\{q_l\}_{1:L}$ as key and value, a cross-attention layer is applied for feature fusion as follow:
\begin{equation}
    \label{eq5}
    \{f^Q_n\}_{1:N} = {Cross\text{-}Att}^{*q}(\{f_n\}_{1:N}, \{q_l\}_{1:L}),
\end{equation}
where $\{f^Q_n\}_{1:N}$ contain the traversal interactions between frame and question token. Our TF is implemented by a $Cross\text{-}Att$ module where $*q$ refers the shared parameters and fusion step with question. 

Another one is to fuse and constrain the temporal representation obtained from T3T by leveraging the text-guided initial video representation. We use $\{f^T_n\}_{1:N}$ as query, $\{f^Q_n\}_{1:N}$ as key and value to obtain final fusion video features:
\begin{equation}
    \label{eq6}
    \{f^C_n\}_{1:N} = {Cross\text{-}Att}^{*t}(\{f^T_n\}_{1:N}, \{f^Q_n\}_{1:N}),
\end{equation}
where the $Cross\text{-}Att$ module is shared and $*t$ is used to distinguish different operation from $*q$. Here, the shared parameters of fusion module can discover and exploit the underlying commonalities of video representations between the $\{f^Q_n\}_{1:N}$ and $\{f^T_n\}_{1:N}$. Finally, TF outputs, computed by cross attention, $\{f^C_n\}_{1:N}$ can be utilized to predict answer.

\subsection{Answer Prediction}

We take a self-attention and cross-attention as Transformer-like structure to predict answer. First, we concatenate $\{f^C_n\}_{1:N}$ and $\{q_l\}_{1:L}$ as a whole and process with a self-attention layer:
\begin{equation}
        H^{Self} = {Self\text{-}Att}([\{f^C_n\}_{1:N}; \{q_l\}_{1:L}]),
\end{equation}
$H^{Self} \in \mathbb{R}^{(N+L) \times D}$ is intermediate variable. $[\cdot;\cdot]$ means splice operation. Finally, the answer candidates $\{a_m\}_{1:M} $ as query, $H^{Self}$ as key and value, the cross-attention layer and Linear layer are adopted to get prediction distribution $\hat{a} \in \mathbb{R}^{ M \times 1}$:
\begin{equation}
        \hat{a} = Linear({Cross\text{-}Att}(\{a_m\}_{1:M}, H^{Self}),
\end{equation}
$Linear(\cdot)$ layer does transformation from $D$ to $1$.

During the training, the ground-truth answer $a^* \in \mathbf{R}^{M \times 1}$ ($0-1$ distribution) as supervision is closed to the prediction by end-to-end optimized with cross entropy loss function:
\begin{equation}
        L = -\sum a^*log(Softmax(\hat{a})).
\end{equation}

\section{Experiments}

\subsection{Datasets and Configuration}
We conduct experiments on three common used VideoQA benchmarks datasets as: NExT-QA~\cite{xiao2021next}, MSVD~\cite{xu2017video} and MSRVTT~\cite{xu2017video}. NExT-QA focuses on temporal and casual reasoning with multi-choice setting, which is mainly used as the main dataset in this paper for extensive ablation validation. MSVD and MSRVTT challenge the description of video with open-ended QA which also include time clues used to demonstrate the universality of our method. We uniformly sample $N\text{=}16$ video frames and extract initial features by pre-trained ViT-L model~\cite{dosovitskiy2021image} which has no longer participated in training. For text, we use a pretrained language model Deberta-base\cite{hedeberta} fine-tuned during the training. The dimensions of the models for features are all set as $D\text{=}768$. Every experiment is done on a single NVIDIA GeForce RTX 4090.


\subsection{Main Results}

As shown in Table~\ref{main}, we compare advanced VideoQA techniques published in recent years. For fair comparison and to explore the temporal learning ability of models, the methods mainly trained with image-based visual features, which can be considered to no temporal information, as input. 

Our method provides a new temporal modeling view to process video feature and has competitive results. The best result on NExT-QA with requirement of complex temporal reasoning ability demonstrates that our method can capture time cues in video and make accurate judgments.
Note that MSVD and MSRVTT focus on description different from NExT-QA where we still obtain best result on MSVD. It reveals that the modeling of time series does not lose the information of visual representation.

\begin{table}[t]
\caption{Accuracy (\%) results on three VideoQA benchmarks datasets. The best is \textbf{bold} and second is \underline{underline}.}
\centering
\vspace{-5pt}
\resizebox{0.90\columnwidth}{!}{
\begin{tabular}{cc|c|c|c}
\hline
    Model & Publish year &  NExT-QA  & MSVD & MSRVTT \\
\hline
    HQGA~\cite{xiao2022aaaivideo} & AAAI22   & 51.8  & 41.2 & 38.6 \\
    KPI(HQGA)~\cite{li2023knowledge} & ICCV23  & 55.0  & 43.3 & 40.0 \\
    VA3(HQGA)~\cite{liao2024align} & CVPR24  & 55.2 & 44.5 & -  \\
\hline 
    MHN~\cite{pengmultilevel} & IJCAI22&  - & 40.4 & 38.6 \\
    EIGV~\cite{li2022equivariant} & MM22&  52.9 & 42.6 & 39.3   \\ 
    VGT~\cite{xiao2022video} & ECCV22  & 53.7 &  -  & 39.7   \\
    VCSR~\cite{wei2023visual} & MM23  & 54.1 & - &  38.9 \\
    PMT~\cite{peng2023efficient} & AAAI23 & - & 41.8 & 40.3  \\
    TIGV~\cite{li2023transformer} & TPAMI23 & 56.7  & 43.1 &  41.1 \\
    PAXION~\cite{wang2024paxion} & NIPS23 & 57.0   & - &  - \\
    MIST~\cite{gao2023mist} & CVPR23 & \underline{57.2}   & - &  - \\
    V-CAT~\cite{zong2024video} & AAAI24 & - & \underline{45.2} &  \textbf{43.3}  \\
\hline   

T3T & (Ours)   & \textbf{61.0} & \textbf{47.3} & \underline{42.9} \\
\hline
\end{tabular}}
\label{main}
\vspace{-5pt}
\end{table}

\subsection{Ablation Study}

\begin{figure}[t]
\centering
\includegraphics[width=0.68\columnwidth]{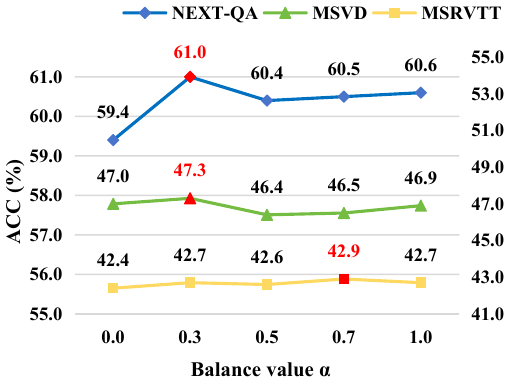}
\vspace{-10pt}
\caption{Comparison of the balance value $\alpha$ for three dataset. We use double axes to distinguish NExT-QA on the left, MSVD and MSRVTT on the right. The best result for each in \textcolor{red}{red}.}
\label{alpha}
\vspace{-15pt}
\end{figure}

\subsubsection{Comparison of balance value \textbf{$\alpha$}}
The core of our T3T implements temporal modeling via TS features and TD features to establish different forms of time characteristic. As the value distribution shown in Fig.~\ref{alpha}, different samples, i.e., data distribution, exhibit varying degrees of sensitivity to the importance of smoothing and difference feature. The balance value $\alpha$ as shown in Fig.~\ref{alpha} is the hyperparameter to adjust the fusion ratio in T3T.
This shows that NExT-QA and MSVD prefer smoothing and successive frames of time clues. On the contrary, it is easier to find answer based on significant difference change on described object for MSRVTT.

\subsubsection{Comparison of components in T3T}

Table~\ref{Components} shows the components in T3T. Compared with Column-1 and Column-2, the model has better performance by TS + TD on NExT-QA than question-guided TF only. However, it is more importance for TF on description-based datasets MSVD and MSRVTT. 
Note that the two temporal modeling approaches, while exhibiting marked differences in their representational orientations, can still exist independently with TF and make meaningful contributions.

\begin{table}[t]
\caption{Comparison of components in T3T. Considering that TS and TD are in parallel, we refer to TS + TD due to series with TF.}
\vspace{-5pt}
\centering
\resizebox{.85\columnwidth}{!}{
\begin{tabular}{l|cccc|ccc}


\hline
                     &1       &2       &3      &4       &5    &6       &7   \\
\hline
  TF                &$\surd$ &        &        &        &$\surd$ &$\surd$ &$\surd$ \\
  TS + TD             &        &$\surd$ &        &        &        &        &$\surd$ \\
   only TS     &        &        &$\surd$        &        &$\surd$ &        &        \\
   only TD     &        &        &        &$\surd$        &        &$\surd$ &        \\
\hline

  NExT-QA            &59.3    &59.7    &59.8       & 50.8      &59.4    &\underline{60.6}    &\textbf{61.0}     \\
  MSVD               &46.7    &46.3    & 45.8     &  32.2    &\underline{47.0}    &46.9    &\textbf{47.3}    \\
  MSRVTT             &42.5    &41.5    & 41.2      &   35.4   &42.4    &\underline{42.7}    &\textbf{42.9}    \\
\hline

\end{tabular}}
\vspace{-10pt}
\label{Components}
\end{table}

\subsection{Further Discussion on Temporal Structure}
In this section, we discuss the design of our modules in constructing temporal features through several comparative experimental settings. To better explore the temporal dynamics, we conduct experiments on dataset NExT-QA, which further divides into three question types of accuracy: Causal~(@C), Temporal~(@T), and Descriptive~(@D).

\subsubsection{Analysis on TF}
Table.~\ref{shared} helps to understand the design for shared cross-attention layer. First, TF provides a strong constraint for the input query. It is informing the model that the input query represents video features. Without TF, the temporal features extracted by the TS + TD will inevitably exhibit a large gap compared to the original features. As shown in Line-5, there is an obvious decline when two independent attention modules are adopted.
Note that *q outperforms *t on @All, but *t exhibits relatively stronger performance on @T where illustrates again the success in temporal reasoning.

\begin{table}[t]
\caption{Comparison of variants for shared cross-attention layer in TF. Improv. is the improvement from \underline{underline} to \textbf{bold}.}
\vspace{-5pt}
\centering
\resizebox{.7\columnwidth}{!}{
\begin{tabular}{cl|ccc|c}
\hline
  & \multirow{2}{*}{Variants}  & \multicolumn{4}{c}{NExT-QA}  \\
  \cline{3-6}
  &                             & @C & @T & @D & @All \\
\hline
  1& T3T & \textbf{59.6}  & \textbf{59.2} & \textbf{68.9} & \textbf{61.0} \\
  2& improv. & +0.9  & +1.5 & +0.3 & +1.1 \\
\hline
  3& only *q & \underline{58.7}  & 57.4 & \underline{68.6} & \underline{59.9}  \\
  4& only *t & 58.1  & \underline{57.7} & 66.6 & 59.3\\
  5& w/o shared & 55.9  & 55.4 & 64.6 & 57.2 \\
\hline
\end{tabular}}
\vspace{-15pt}
\label{shared}
\end{table}

\subsubsection{Analysis on TS}
In TS module, we use the definition of Brown Bridge process to model smooth video representations and $ConV(\cdot)$ layers are set as stochastic part. When processing video features with high information redundancy, dropout is often employed as a common used technique to mitigate the risk of overfitting. However, there is not a required component shallow in convolution layers as shown in Table.~\ref{Smoothing}. From video representation to time series, this process can be understood as a form of degradation focusing on the temporal changes without overfitting. Compared to Line-1 with Line-3, accuracy of @T increases $1.7\%$, which demonstrates that the model is more robust to temporal learning and this method still has unexplored potential.

\begin{table}[t]
\caption{Comparison of variants for the Brown Bridge structure in TS. K stands for the layer number of $ConV(\cdot)$. Drop. increases a dropout layer at each end of $ConV(\cdot)$.}
\vspace{-5pt}
\centering
\resizebox{.7\columnwidth}{!}{
\begin{tabular}{cl|ccc|c}
\hline
  &\multirow{2}{*}{Variants}  & \multicolumn{4}{c}{NExT-QA}  \\
  \cline{3-6}
   &                           & @C & @T & @D & @All \\
\hline
   1& K=2 (T3T)     & \textbf{59.6}  & \textbf{59.2} & \textbf{68.9} & \textbf{61.0} \\
   2& K=2 + Drop. & 58.0  & \underline{58.5} & 66.6 & 59.6  \\
   3& K=1           & \underline{59.3}  & 57.5 & \underline{68.3} & \underline{60.2}  \\
   4& K=1 + Drop. & 58.5  & 57.0 & 68.0 & 59.6  \\
\hline
\end{tabular}}
\label{Smoothing}
\vspace{-10pt}
\end{table}

\subsubsection{Analysis on TD}

TD as the module without trainable parameters focuses on the most pure change of video features. Even if the two adjacent sampled frames already have a certain time span. Compared Line-3 with Line-1 as shown in Table.~\ref{Difference}, sampling frames by interval may aggravate the variation of the features resulting in poorer performance. Moreover, Softmax operation provides a set of values to enhance the high amplitude location that prompts the model to pay stronger attention. Increasing interval $I$ is better without softmax operation as shown in Line-4. When frames interval is too far, the enhancement of the amplitude is not required.

\begin{table}[t]
\caption{Comparison of variants for the Difference structure in TD. Line-2 and Line-4 delete the Softmax operation.}
\vspace{-5pt}
\centering
\resizebox{.8\columnwidth}{!}{
\begin{tabular}{cl|ccc|c}
\hline
  &\multirow{2}{*}{Variants}  & \multicolumn{4}{c}{NExT-QA}  \\
  \cline{3-6}
   &                           & @C & @T & @D & @All \\
\hline
   1& I=0 (T3T)    & \textbf{59.6}  & \textbf{59.2} & \textbf{68.9} & \textbf{61.0} \\
   2& I=0 w/o Soft. & 58.9  & 58.2 & 67.8 & 60.1  \\
   3& I=1           & 57.4  & 57.4 & \underline{68.5} & 59.2  \\
   4& I=1 w/o Soft. & \underline{59.0}  & \underline{58.6} & 68.0 & \underline{60.4}  \\
\hline
\end{tabular}}
\label{Difference}
\vspace{-15pt}
\end{table}

In summary, the above three sets of experimental results indicate significant importance and interpretability for temporal modeling on long-form video NExT-QA. This provides new perspective on the extraction and utilization of temporal representations when modeling video features for VideoQA.

\vspace{-3pt}
\subsection{Visualization}

\begin{figure}[t]
\centering
\includegraphics[width=1.0\columnwidth]{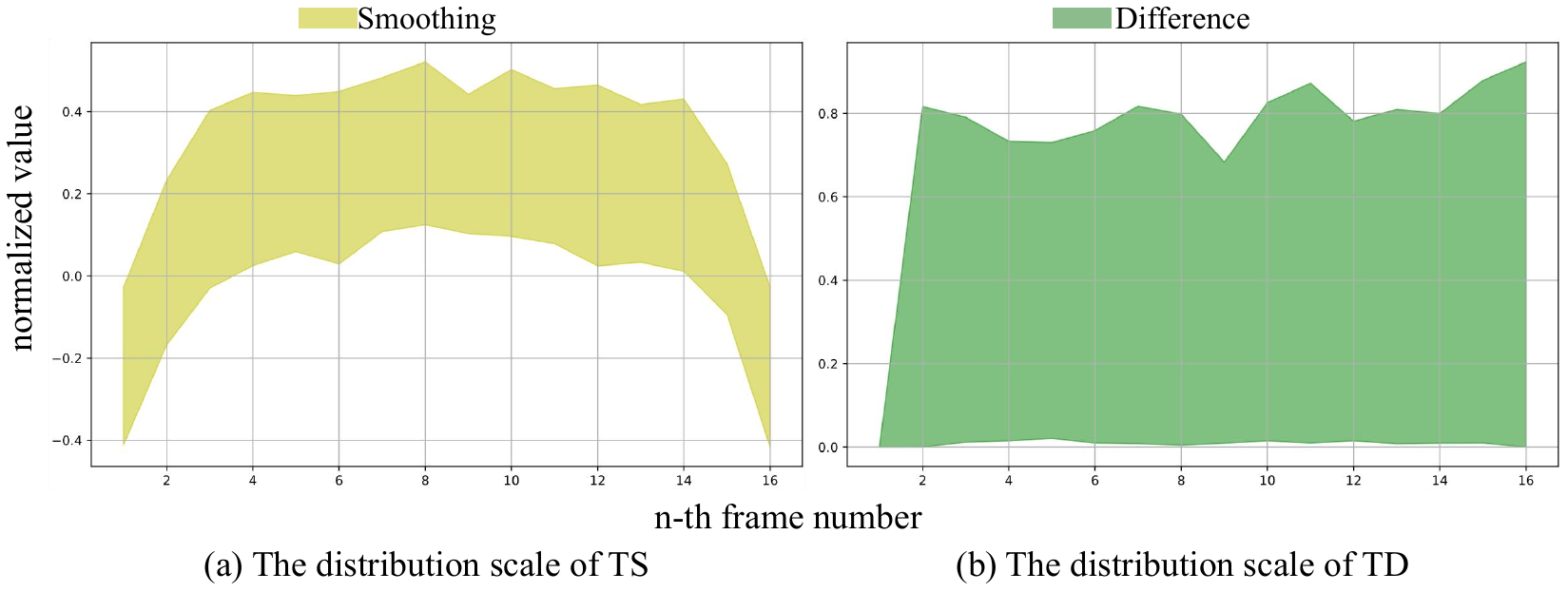} 
\vspace{-15pt}
\caption{The normalized distribution scale extracted by TS module in \textcolor{deepyellow}{yellow} and TD module in \textcolor{deepgreen}{green} of the whole NExT-QA test set.}
\vspace{-15pt}
\label{distribution}
\end{figure}

The Fig.~\ref{distribution} show the distribution scales on the NExT-QA test set, which are separately extracted and normalized from TS and TD.
For TS, the start and end are defined as initial frames features with certainty and the stochastic part is learnable with uncertainty by Brownian Bridge constraint.
As shown in Fig.~\ref{distribution}~(a), the distribution is more elongated, and it indeed exhibits the shape of a bridge. 
By contrast for TD, the start is defined as zero and difference operation is applied to two adjacent frames. For different samples, the salience magnitude of this change distribution does not follow a pattern just relying on the video itself. Thus the distribution scale is uniform as shown in Fig.~\ref{distribution}~(b). 
The distinct scales of the two distributions prominently highlight more comprehensive and interpretable temporal modeling. 

\vspace{-5pt}
\section{Conclusion}
\vspace{-3pt}
In this paper, we introduced a novel framework called Temporal Trio Transformer (T3T) designed for the VideoQA task. Our approach leverages both the smoothing dynamics and local difference within video content to explore temporal consistency and variability. Temporal Smoothing (TS) based on Brownian Bridge Process and Temporal Difference (TD) by computing frame differences collaboratively establish multi-temporal representations. Moreover, Temporal Fusion (TF) effectively integrates text-guided nuanced understanding of the video in relation to the questions. Extensive experimental validation on multiple datasets demonstrates the superiority of our T3T framework in handling the complexities of VideoQA. Our work provides a new view for video modeling and reveals the immense potential of temporal learning, with the hope of driving further research in video understanding and representation.

\vspace{-3pt}

\bibliographystyle{IEEEbib}
\vspace{-5pt}
\bibliography{VideoQAreferences}

\end{document}